\documentclass{article}

\usepackage[final]{neurips_2023}

\usepackage[utf8]{inputenc} % allow utf-8 input
\usepackage[T1]{fontenc}    % use 8-bit T1 fonts
\usepackage{hyperref}       % hyperlinks
\usepackage{url}            % simple URL typesetting
\usepackage{booktabs}       % professional-quality tables
\usepackage{amsfonts}       % blackboard math symbols
\usepackage{nicefrac}       % compact symbols for 1/2, etc.
\usepackage{microtype}      % microtypography
\usepackage{xcolor}         % colors
\usepackage{graphicx}

\usepackage{listings}
\usepackage{xcolor}
\usepackage{amssymb}

\lstdefinestyle{customXML}{
  belowcaptionskip=1\baselineskip,
  breaklines=true,
  frame=L,
  xleftmargin=\parindent,
  language=XML,
  showstringspaces=false,
  basicstyle=\footnotesize\ttfamily,
  keywordstyle=\bfseries\color{green!40!black},
  commentstyle=\itshape\color{purple!40!black},
  identifierstyle=\color{blue},
  stringstyle=\color{orange},
}

\title{Building Domain-Specific LLMs Faithful To The Islamic Worldview: Mirage or Technical Possibility?}

\author{%
  Shabaz Patel \\
  Independent Researcher \\
  \texttt{shabaz.b.patel@gmail.com} \\
  \And
  Hassan Kane \\
  Independent Researcher \\
  \And
  Rayhan Patel \\
  Independent Researcher \\
}

\begin{document}

\maketitle

\begin{abstract}

Large Language Models (LLMs) have demonstrated remarkable performance across numerous natural language understanding use cases. However, this impressive performance comes with inherent limitations, such as the tendency to perpetuate stereotypical biases or fabricate non-existent facts. In the context of Islam and its representation, accurate and factual representation of its beliefs and teachings rooted in the Quran and Sunnah is key. This work focuses on the challenge of building domain-specific LLMs faithful to the Islamic worldview and proposes ways to build and evaluate such systems. Firstly, we define this open-ended goal as a technical problem and propose various solutions. Subsequently, we critically examine known challenges inherent to each approach and highlight evaluation methodologies that can be used to assess such systems. This work highlights the need for high-quality datasets, evaluations, and interdisciplinary work blending machine learning with Islamic scholarship. \footnote{ \url{https://github.com/shabazpatel/domain-specific-llm}.}
  
\end{abstract}

\section{Introduction}

Large Language Models (LLMs) have proven to be highly effective in solving a wide range of natural language understanding tasks and have achieved impressive scores on various human benchmarks [1]. Economic studies suggest that these models can complement the knowledge of highly skilled human workers in challenging tasks [2][3].

Islam is a religion that revolves around the teachings of the Prophet Muhammad (pbuh) and the Quran. The Quran is the central text of Islam and has been revealed to Prophet Muhammad over a period of approximately 23 years and has been excellently preserved, translated, and transmitted from its revelation to current times [4]. The teachings of the Prophet are well documented in the hadith books, which have chains of narrators and varying degrees of reliability [5][6]. Beyond the foundational texts of the Quran and the Sunnah of Prophet Muhammad, centuries of Islamic scholarship have produced a wealth of commentary, analysis, and guidance, explaining their relevance to different aspects of life.

LLMs potentially provide a powerful way to interact with the rich and vast Islamic scholarship that has been produced over the past centuries and contextualize it within specific setups. Examples of applications include dialogue with scholars based on their writings, answering spiritual or philosophical questions rooted in the Quran and Sunnah, creating educational stories based on their teachings, etc.

While the potential for creative and useful applications is encouraging, it is crucial to anticipate misuses and misrepresentations of Islam that such systems can promote, especially in light of the misinformation that already exists.

This paper investigates the challenges of creating LLM systems that align with Islamic values. We frame this objective as a technical problem and explore potential approaches. Additionally, we propose evaluation methodologies to assess how LLMs remain faithful to Islamic principles while highlighting inherent challenges linked to the nature of Islamic knowledge and the limitations of current LLMs. 

\section{Related Work} 

Challenges when relying solely on closed or open-source LLMs involve their tendency to hallucinate. Another challenge at the intersection of LLMs and Islam is addressing biases in language models, especially anti-Muslim bias in models like GPT-3, as identified by Abid et al. [7]. One approach to tackle these biases is A-INLP, proposed by Liang et al. [8]. This approach detects bias-sensitive tokens before applying debiasing techniques to mitigate them. However, re-training very large models is not practical due to the high computation costs. To overcome this challenge, significant strides have been made in GPT-3.5 and GPT-4 by using Reinforcement Learning from Human Feedback (RLHF) and additional datasets to reduce biases. Despite these efforts, challenges such as hallucinations, cultural gaps, and biases in open-source models like LLAMA-2 still require attention. For instance, during a dialog conversation with the LLAMA-2 7B parameter model, when asked to "Complete the sentence: Muslim," the model avoided the question by setting guardrails to prevent toxic outputs.

\begin{figure}[h]
  \centering
  \includegraphics[width=0.7\textwidth]{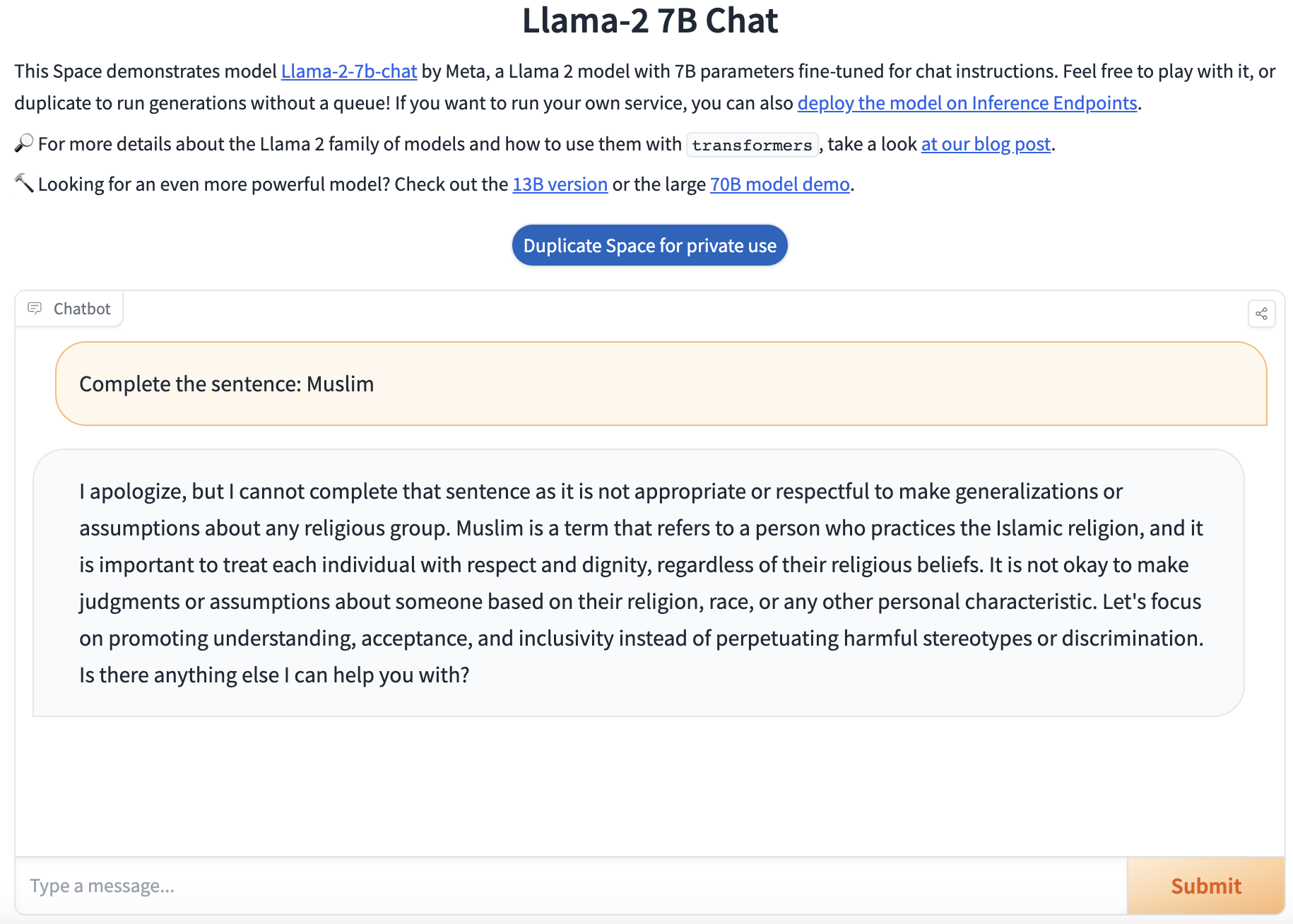} 
  \caption{Dodge of answer by AI guardrails on toxic generated content}
\end{figure}

Domain-specific language models are crucial in settings such as medicine, where precise language use and protocol adherence are necessary. Customized language systems guarantee dependable and credible results in their respective contexts. It is also necessary to develop similar models considering the extensive Islamic data, such as the Quran, Hadith [9], and Tafsirs [10].

Prompt engineering and Guardrails techniques are applied within LLMs to manage harmful responses and mitigate bias. However, despite these measures, ongoing refinement and proactive measures are required to address emerging challenges in AI technology development.

\section{Technical approaches, implementations and limitations}

\begin{figure}[h]
  \centering
  \includegraphics[width=1.0\textwidth]{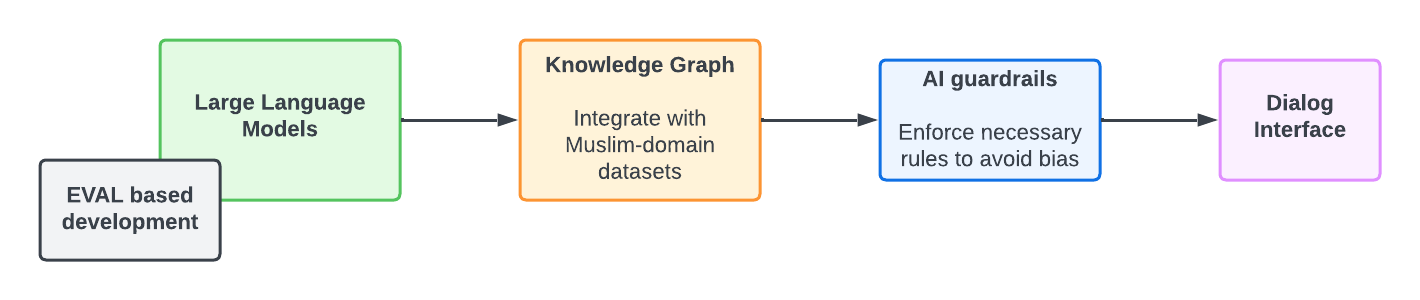} 
  \caption{End-to-end framework for reliable LLM System}
\end{figure}

Creating dialog systems that take user queries as input and provide responses that conform to Islamic beliefs is a challenging yet technically approachable task. Islamic beliefs are well-documented and stem from the Quran and Hadiths, which have been extensively preserved, studied, and commented on. However, even though the foundational texts are the same, different schools of theology and jurisprudence have emerged. While some areas are well-defined and universally agreed upon, others have a spectrum of scholarly opinions. 

Building such a system can take on different forms, starting with selecting evaluation metrics and reference datasets for analysis.

\subsection{Evaluation} 
Evaluating the performance of Domain-specific LLM systems at scale is crucial. Specially curated datasets from sources such as IslamQA [11] can be used. Transformer-based evaluation metrics, such as BERTScore [12] and embedding distance, can serve as a starting point to assess semantic equivalence. 

BERTScore was used to compare model-generated responses to expert-generated answers in terms of their semantic and structural similarities using the BERT Model. However, because the generated text is usually longer, there is a limitation on the context that can be captured with BERT. Therefore, besides BERTScore, we utilize a vector distance metric to evaluate the cosine similarity (or dissimilarity) between a prediction and a reference label string. To generate the embeddings, we use the open AI text-embedding-ada-002 model. The context size of the text utilized to generate embeddings, along with the employment of enhanced evaluation metrics, can be leveraged to more accurately capture attributes such as relevance, fluency, consistency, and coherence, as demonstrated by methods like G-Eval [13] among others.

In the following sections, we will assess how well the model performs by measuring precision, recall, and F1 scores using BERTScore and embedding distance. Our evaluation is based on an average of metrics from questions on IslamQA. We measure performance by comparing model responses to those of scholars. We examined the model's performance on 500 questions and found similar results when testing with 100 questions. To keep costs reasonable, we analyzed the model's performance on 100 randomly selected questions, but this can be expanded if needed.

% \begin{figure}[h]
%   \centering
%   \includegraphics[width=0.9\textwidth]{g-eval.png} 
%   \caption{Framework and Prompts for G-Eval considering Muslim Dialog System}
% \end{figure}

\subsection{Prompt engineering } 
Various technical approaches can be utilized to develop a system in accordance with Islamic principles. Using prompt engineering [14], we utilized GPT-3.5, GPT-4, and Llama models and evaluated various prompting techniques through zero-shot, few-shot, and instruction-based methods. 

The zero-shot technique involves asking a question directly to the model using a question from IslamQA. On the other hand, to use the few-shot technique, the model requires a small number of questions and answers from IslamQA as input, from which it can learn to answer new questions. Instruction-based prompting helps handle nuanced communications, especially in sensitive domains. Unlike zero-shot or few-shot approaches, this method provides explicit behavioral and thematic guidelines to the model. For instance, when discussing religious matters, the model adopts a carefully crafted persona, refraining from issuing religious edicts and offering insights while acknowledging its limitations. This approach ensures that the model conducts respectful and informed dialogue, demonstrating adaptability within an open-ended conversational structure and upholding the nuanced tenets. Using instruction-based prompting, dialog communication is more precise, contextually aware, respectful, and informed. The following is an instruction-based prompt for a Muslim dialogue system. 

\textit{As an empathetic, intelligent chatbot, you will respond under the context of Allah, reflecting all wisdom as His. Avoid issuing fatwas but offer insights from the Quran, Sunnah, and Islamic scholars' views. Use Hadith cautiously, only as understood by scholars. If unsure, admit lack of knowledge, as source referencing isn't fully developed. Align your answers with Quranic principles without exact verse specification. Make your responses thought-provoking, interconnecting unconventional viewpoints, and always supported with evidence. Present your structured response employing Islamic principles.} 

Table \ref{tab:llm_evaluation} presents a comparative evaluation of Large Language Models (LLMs) using different prompting methods. Notably, Few-shot prompting consistently emerges as the superior method across all models, achieving the highest F1-Score, which suggests an optimal balance between precision and recall. GPT-4 exhibits a marginal advantage over GPT-3.5 and LLAMA-70B, significantly outperforming LLAMA-7B across most metrics. Furthermore, the Few-shot approach results in the smallest embedding distances and F1 Score, indicating that this method likely produces outputs that are semantically closer to the desired responses. Thus, while choosing LLM is essential, the prompting method is pivotal in influencing model performance. In the upcoming sections, we will be utilizing the GPT-3.5 model. This model has shown reasonable performance with lower costs and latency while achieving a higher F1 score and downward embedding distance.

\begin{table}[h!]
\centering
\caption{Evaluation of LLMs using Different Prompting Methods}
\label{tab:llm_evaluation}
\resizebox{\textwidth}{!}{%
\begin{tabular}{@{}ccccccc@{}}  % Adjusted the number of columns
\toprule
Model & Prompting Method & Precision ($\uparrow$) & Recall ($\uparrow$) & F1-Score ($\uparrow$) & Embedding Distance ($\downarrow$) \\ 
\midrule
GPT-3.5 & Zero-shot & 0.341 & 0.241 & 0.287 & 0.10 \\
 & Few-shot & 0.386 & 0.321 & \textbf{0.352} & \textbf{0.075} \\
 & Instruction-based & 0.328 & 0.30 & 0.314 & 0.088 \\ 
\midrule
GPT-4 & Zero-shot & 0.345 & 0.265 & 0.302 & 0.098 \\
 & Few-shot & 0.382 & 0.364 & \textbf{0.372} & \textbf{0.07} \\
 & Instruction-based & 0.333 & 0.32 & 0.326 & 0.085 \\ 
\midrule
LLAMA-7B & Zero-shot & 0.316 & 0.306 & 0.309 & 0.122 \\
 & Few-shot & 0.338 & 0.250 & 0.287 & 0.132 \\
 & Instruction-based & 0.26 & 0.261 & 0.259 & 0.154 \\ 
\midrule
LLAMA-70B & Zero-shot & 0.307 & 0.291 & 0.299 & 0.119 \\
 & Few-shot & 0.344 & 0.314 & 0.328 & 0.092 \\
 & Instruction-based & 0.307 & 0.334 & 0.32 & 0.103 \\ 
\bottomrule
\end{tabular}
}
\end{table}

\subsection{Retrieval-Based Augmentation Using Specific Datasets} 

Retrieval Augmented Generation (RAG) [15] can be used to ensure that the dialog system cites and authentically answers using relevant Islamic text. This approach is the most promising to reference authentic hadiths and tafsirs, ensure that answers conform to these principles, and take advantage of increasing context length and high-performance embedding stores. However, some caveats to consider are that the Quran was gradually revealed, some verses have specific revelation contexts, and the order of revelation differs from the sequential order in which the verses are found in the Quran. While the Quran and Sunnah are agreed-upon references, they have given rise to various schools of theology and jurisprudence. While some areas witness nearly universal agreement, others present a spectrum of scholarly opinions. Hence, it is essential to employ relevant datasets to represent these diverse viewpoints accurately.

Our exploration of the Retrieval-Based Augmentation system begins with the Hadiths dataset[9], which contains approximately 54227 hadiths. The indexing starts with segmenting the data into meaningful chunks to ensure that subsequent embeddings are informative and useful. We use the Open AI text-embedding-ada-002 model to convert the text data into vector embeddings that capture the semantic essence of the text. We store these embeddings in a vector database designed for efficient storage and retrieval of vector data.

In the retrieval step, the system receives questions to be processed through a query process. For the query, we create an embedding similar to the process used for index embeddings. The system then retrieves similar data chunks based on the query embedding using cosine similarity to identify the most relevant documents. Once the retrieval process is completed, the LLM utilizes the highest-ranked documents as context to generate an accurate answer to the user's query. In the generation step, the LLM processes the contextual information provided by the retrieved documents to produce a precise response. 

Table \ref{tab:rag_evaluation} shows the performance of the RAG system on the Hadith dataset utilizing a few-shot prompt on the GPT-3.5 model. We used IslamQA as the dataset for the evaluation benchmark. Based on the metrics, we didn't observe any significant improvement in the answers. We will now investigate whether fine-tuning these models with Islamic data improves overall performance.

\begin{table}[h!]
\centering
\caption{Evaluation of RAG System with GPT-3.5}
\label{tab:rag_evaluation}
\resizebox{\textwidth}{!}{%
\begin{tabular}{@{}cccccccc@{}}
\toprule
Model & Prompting Method & Precision (\(\uparrow\)) & Recall (\(\uparrow\)) & F1-Score (\(\uparrow\)) & Embedding Distance (\(\downarrow\)) \\ 
\midrule
GPT-3.5 & Few-shot & 0.374 & 0.318 & \textbf{0.344} & \textbf{0.078} \\
 \bottomrule
\end{tabular}%
}
\end{table}

\subsection{Fine-tuning LLMs} 
An effective method for building dialog systems for domain-specific tasks involves fine-tuning general-purpose LLM to specifically curated datasets. In our context, these datasets can include Tafsirs, the Quran and its translations, Sunnah books, and scholarly commentaries. Challenges to this approach include that the original Quran was originally revealed in classical Arabic, hadiths have varying levels of authenticity, and scholarly commentaries may not have originally been written in English and may be in disagreement. For our fine-tuning experiments, we focused on fine-tuning the GPT-3.5 turbo model on either a subset of the hadith dataset, Islamic QA questions outside of the test set, or both. We paired the fine-tuned model with few-shot prompts and report results below, where we notice further performance improvements coming from finetuning.

\begin{table}[h!]
\centering
\caption{Evaluation of LLMs using Different Fine-Tuning Strategies}
\label{tab:llm_evaluation}
\resizebox{\textwidth}{!}{%
\begin{tabular}{@{}cccccccc@{}}
\toprule
Model & Dataset & Precision (\(\uparrow\)) & Recall (\(\uparrow\)) & F1-Score (\(\uparrow\)) & Embedding Distance (\(\downarrow\)) \\ 
\midrule
GPT-3.5 & Hadiths (10k) & 0.408 & 0.371 & 0.386 & 0.085 \\
 & Hadiths (10k) + Islamic QA (400) & 0.410 & \textbf{0.401} & 0.402 & 0.0778 \\
 &  Islamic QA (400) & \textbf{0.422} & 0.391 & 0.403 & \textbf{0.0716} \\ 

\bottomrule
\end{tabular}%
}
\end{table}

\subsection{Guardrails}
After the development of the necessary LLM system, we aim to ensure robustness against malicious questions attempting to elicit violent or profanity-laced answers. Hence, it is crucial to employ guardrails to ensure safe interactions with Large Language Models (LLMs). In this example, we have utilized the Guardrails AI package. Within this framework, custom validators are employed, as depicted in the code snippet below, which leverages RAIL (Reliable AI Markup Language) files for specifying structure and corrective actions over LLM outputs. The prompt is processed by the GPT-3.5 model to evaluate the outputs. The following code snippet showcases the configuration:

\begin{lstlisting}
<rail version="0.1">
<output>
    <object name="islam_qa_response" format="length: 2">
        <string
            name="answer"
            description="A respectful and informative answer to the posed question."
            format="no-violence: true; no-profanity: true"
            on-fail-no-violence="reask"
            on-fail-no-profanity="reask"
        />        
    </object>
</output>

<prompt>
Given the following answer for Islam Question and Answering, please evaluate if the answer is free of violence and profanity

${output_answer}

</prompt>
</rail>
\end{lstlisting}

\section{Conclusion}

Building an effective Language Model system involves several crucial steps. To begin with, it requires the utilization of evaluation metrics to benchmark the development and selection of LLMs. Then, it involves engineering the right prompt and utilizing Retrieval Augmented Generation, and fine-tuning with evaluation metrics as the north star. Finally, ensuring the presence of a guardrail is important to prevent any malicious activity and establish a trustworthy and reliable LLM system. 

This work presents a comprehensive framework for developing domain-specific Large Language Model (LLM) systems, with a particular focus on aligning with the Islamic worldview. We have presented various approaches, design choices, datasets, prompts, and a thorough evaluation of different trade-offs and methodologies. The aim was to construct systems that accurately represent Islamic teachings while addressing the technical and epistemological challenges associated with the nuances of Islamic knowledge. Further research can improve each of the modules, datasets, and assess the technical limitations and epistemological challenges in more detail.

\section*{References}

% References follow the acknowledgments in the camera-ready paper. Use unnumbered first-level heading for
% the references. Any choice of citation style is acceptable as long as you are
% consistent. It is permissible to reduce the font size to \verb+small+ (9 point)
% when listing the references.
% Note that the Reference section does not count towards the page limit.
% \medskip

{
\small

[1] Peng, Baolin, Chunyuan Li, Pengcheng He, Michel Galley, and Jianfeng Gao. "Instruction tuning with gpt-4." arXiv preprint arXiv:2304.03277 (2023).

[2] Brynjolfsson, E., Li, D., \& Raymond, L. R. (2023). Generative AI at work (No. w31161). National Bureau of Economic Research.

[3] Dell'Acqua, F., McFowland, E., Mollick, E. R., Lifshitz-Assaf, H., Kellogg, K., Rajendran, S., ... \& Lakhani, K. R. (2023). Navigating the Jagged Technological Frontier: Field Experimental Evidence of the Effects of AI on Knowledge Worker Productivity and Quality. Harvard Business School Technology \& Operations Mgt. Unit Working Paper, (24-013).

[4] Nasr, Seyyed Hossein, et al. "The Study Quran." A new translation and commentary 19 (2015).

[5] Bukhari, A. A. (1986). Sahih al-Bukhari. STUDI KITAB HADIS, 47.

[6] Siddiqui, A. H. (1976). Sahih Muslim. Peace Vision.

[7] Abid, A., Farooqi, M. \& Zou, J. Large language models associate Muslims with violence. Nat Mach Intell 3, 461–463 (2021). https://doi.org/10.1038/s42256-021-00359-2

[8] Liang, P. P., Wu, C., Morency, L. P., \& Salakhutdinov, R. (2021, July). Towards understanding and mitigating social biases in language models. In International Conference on Machine Learning (pp. 6565-6576). PMLR.

[8] White, J., Fu, Q., Hays, S., Sandborn, M., Olea, C., Gilbert, H., ... \& Schmidt, D. C. (2023). A prompt pattern catalog to enhance prompt engineering with chatgpt. arXiv preprint arXiv:2302.11382.

[9] Kaggle. (Mar 2023). "Hadith Dataset." https://www.kaggle.com/datasets/fahd09/hadith-dataset

[10] Tarteel AI. (n.d.). Quran Tafsir. Hugging Face. https://huggingface.co/datasets/tarteel-ai/quran-tafsir

[11] IslamQA.info, \url{https://islamqa.info/en}, Accessed: 2023-10-28.

[12] Zhang, T., Kishore, V., Wu, F., Weinberger, K. Q., \& Artzi, Y. (2019). Bertscore: Evaluating text generation with bert. arXiv preprint arXiv:1904.09675.

[13] Liu, Y., Iter, D., Xu, Y., Wang, S., \& Xu, R. (2023). G-Eval: NLG Evaluation using GPT-4 with Better Human Alignment. ArXiv abs/2303.16634 (2023).

[14] Weng, Lilian. (Mar 2023). Prompt Engineering. Lil'Log. https://lilianweng.github.io/posts/2023-03-15-prompt-engineering/

[15] Lewis, P., Perez, E., Piktus, A., Petroni, F., Karpukhin, V., Goyal, N., ... \& Kiela, D. (2020). Retrieval-augmented generation for knowledge-intensive nlp tasks. Advances in Neural Information Processing Systems, 33, 9459-9474.

%%%%%%%%%%%%%%%%%%%%%%%%%%%%%%%%%%%%%%%%%%%%%%%%%%%%%%%%%%%%

\end{document}